\documentclass{article}

\usepackage[english]{babel}
\usepackage{multirow}
\usepackage{graphicx}

\usepackage[letterpaper,top=2cm,bottom=2cm,left=3cm,right=3cm,marginparwidth=1.75cm]{geometry}

\usepackage{amsmath}
\usepackage{graphicx}
\usepackage{colortbl}
\usepackage{xcolor}
\usepackage{pifont}
\usepackage{authblk}

\makeatletter
\newcommand{\printfnsymbol}[1]{%
  \textsuperscript{\@fnsymbol{#1}}%
}
\makeatother
\usepackage[colorlinks=true, allcolors=blue]{hyperref}
\newlength\savewidth\newcommand\shline{\noalign{\global\savewidth\arrayrulewidth\global\arrayrulewidth1pt}\hline\noalign{\global\arrayrulewidth\savewidth}}
\definecolor{baselinecolor}{gray}{.9}
\definecolor{darkgray}{gray}{.7}
\newcommand{\cb}{\cellcolor{green!30}}

\title{{\bf Gene42}\footnote{Gene42 is part of the Omics42 platform which includes also a family of protein LMs and a family of chemical LMs named Prot42 and Chem42, respectively. Refer to \href{https://huggingface.co/spaces/inceptionai/Omics42}{\textbf{Omics42}} blog at \href{https://huggingface.co/inceptionai}{huggingface.co/inceptionai} for further details.}: Long-Range Genomic Foundation Model With Dense Attention}

\author[1]{Kirill Vishniakov\thanks{These authors contributed equally to this work.}}
\author[2]{Boulbaba {Ben Amor}\printfnsymbol{2}\thanks{Corresponding author: Boulbaba Ben Amor \url{boulbaba.amor@inceptionai.ai}}}
\author[3]{Engin Tekin}
\author[2]{Nancy A. ElNaker}
\author[1]{Karthik Viswanathan}
\author[1]{Aleksandr Medvedev}
\author[2]{Aahan Singh}
\author[2]{Maryam Nadeem}
\author[2]{Mohammad Amaan Sayeed}
\author[1]{Praveenkumar Kanithi}
\author[1]{Tiago Magalhaes}
\author[3]{Natalia Vassilieva}
\author[2]{Dwarikanath Mahapatra}
\author[1]{Marco Pimentel}
\author[1]{Shadab Khan}

\affil[1]{M42, Abu Dhabi, UAE.}
\affil[2]{Inception Institute of Artificial Intelligence, Abu Dhabi, UAE.}
\affil[3]{Cerebras Systems, Sunnyvale, CA, USA.}

\date{}

\begin{document}
\maketitle

\begin{abstract}

We introduce Gene42, a novel family of Genomic Foundation Models (GFMs) designed to manage context lengths of up to 192,000 base pairs (bp) at a single-nucleotide resolution. Gene42 models utilize a decoder-only (LLaMA-style) architecture with a dense self-attention mechanism. Initially trained on fixed-length sequences of 4,096 bp, our models underwent continuous pretraining to extend the context length to 192,000 bp. This iterative extension allowed for the comprehensive processing of large-scale genomic data and the capture of intricate patterns and dependencies within the human genome. Gene42 is the first dense attention model capable of handling such extensive long context lengths in genomics, challenging state-space models that often rely on convolutional operators among other mechanisms. Our pretrained models exhibit notably low perplexity values and high reconstruction accuracy, highlighting their strong ability to model genomic data. Extensive experiments on various genomic benchmarks have demonstrated state-of-the-art performance across multiple tasks, including biotype classification, regulatory region identification, chromatin profiling prediction, variant pathogenicity prediction, and species classification. The models are publicly available at \href{https://huggingface.co/inceptionai}{huggingface.co/inceptionai}.

\end{abstract}

\section{Introduction}
\label{sec:Introduction}

Understanding the intricate details of the human genome is crucial for advancing our comprehension of diseases and developing personalized therapies tailored to an individual's genetic makeup. This approach improves treatment efficacy and minimizes adverse effects~\cite{brittain2017rise}, marking a significant shift from the traditional "one-size-fits-all" model of medicine to a more targeted and effective strategy. The Human Genome Project, completed in 2003, marked a pivotal moment in genomic analysis. It provided a comprehensive reference that revolutionized the understanding of human genetics and laid the foundation for modern genomic research. Since then, advances in DNA sequencing technologies have significantly reduced the cost and time required to sequence genomes, improving access to these technologies and enabling precision medicine initiatives around the world, while fueling innovation in related fields~\cite{satam2023next}.

Building foundation models for genomic data analysis has been a dynamic area of research~\cite{consens2023transformers}, producing models that are evaluated on various downstream tasks and benchmarks~\cite{grevsova2023genomic, liu2024genbench, dalla2023nucleotide}. These models, referred to as Genomic Foundation Models (GFMs), leverage a range of advanced architectures and deep learning techniques to decode the complex language of DNA. By capturing long-range dependencies and contextual information within genomic sequences, they aim to provide more precise predictions and insights into genomic functions and interactions between different elements of the genome~\cite{consens2023transformers}. GFMs are designed to take gene sequencing data as input to facilitate tasks such as predicting gene expression, identifying regulatory elements, and modeling complex biological systems. By improving the understanding of genomic sequences, these models can accelerate discoveries in precision medicine and drug design, leading to more effective solutions.

The importance of such models is highlighted by two aspects of genomic data that have influenced the development of GFMs:
\textbf{(i)} Single Nucleotide Polymorphisms (SNPs): Changes in a single nucleotide can significantly alter genes by affecting the coding sequence or regulatory elements. These changes can impact protein function or expression levels, potentially leading to the production of non-functional proteins, alterations in the protein activity or stability, or the activation of disease pathways. Such alterations can cause cells to enter disease states, including cancers, genetic disorders, and other health conditions~\cite{shastry2009snps}.
\textbf{(ii)} Long-range Genomic Interactions: Genomic data exhibits long-range interactions, where regions of the genome that are distant in the linear sequence can physically interact through the three-dimensional folding of DNA. These interactions are crucial for regulating gene expression, as enhancers and other regulatory elements can influence genes located mega-bases away. Understanding these interactions is essential for comprehending complex regulatory networks that control cellular functions and how disruption of these networks can lead to diseases~\cite{sanyal2012long}.\\

Based on the above analysis, in this paper, we present a new family of Genomic Foundation Models capable of handling medium and long-range context lengths.  We adopt a decoder-only LLaMA-style architecture, pretrained on genomic sequences. We have built a state-of-the-art (SoTA) family of transformer-based models leveraging dense attention for robust contextual understanding and the ability to capture long-range (up to 192 kbp) dependencies at single-nucleotide resolution, which represents the main contribution of this work. We demonstrate the efficiency of our model family across various short, medium, and long-range genomic tasks. The rest of the paper is structured as follows. In Section \ref{sec:SoTA}, we begin with a brief literature review on Genomic Foundation Models. Section \ref{sec:Methodology} outlines our methodology, detailing the process from genomic data tokenization to the development of our models with extended perceptive field through incremental continuous pretraining. Section \ref{sec:experiments} presents our comprehensive evaluations and comparisons to SoTA models. Finally, Section \ref{sec:Conclusion} offers concluding remarks and discusses the implications of our work for genomic data analysis and related challenges.

\section{Related Works}
\label{sec:SoTA}

\textbf{BERT-based approaches.} Foundation models for genomic data have seen substantial progress with the introduction of DNABERT \cite{ji2021dnabert} and its successor, DNABERT-2 \cite{zhou2023dnabert}, which use BERT (Bidirectional Encoder Representations from Transformers) architecture. DNABERT's approach of treating k-mers as tokens and utilizing bidirectional transformers enabled it to achieve high accuracy in various genomic prediction tasks. k-mer tokenization breaks the sequence into fixed-length subsequences, which can result in the loss of important contextual information that spans beyond the k-mer length. For instance, motifs or regulatory elements longer than k might be split across multiple k-mers. DNABERT-2, in contrast, builds upon its predecessor by employing an advanced tokenization strategy called Byte Pair Encoding (BPE) \cite{zhou2023dnabert}. The Nucleotide Transformer (NT) \cite{dalla2023nucleotide} combines a BERT-based encoder and a k-mer tokenization strategy. Trained on diverse datasets, including the Human Reference Genome (HRG), 1000 Genomes, and multispecies genome, NT models capture transferable, context-specific representations of k-mer sequences. The NT work also established a new benchmark called the Nucleotide Transformer (NT) Benchmark and a leaderboard to evaluate models on 18 genomic prediction tasks. GENA-LM \cite{fishman2023gena}, another family of models following a similar approach, uses a BERT architecture and includes a family of transformer-based models designed to process extensive DNA sequences up to 36k base pairs. Utilizing BPE tokenization, GENA-LM can encode long sequence fragments. In addition to the Human Reference Genome (HRG), GENA-LM uses multispecies data and variations curated from the 1000G Project. More recently, \cite{kuratov2024recurrent} investigated enhancing GENA-LM models using Recurrent Memory Transformers (RMT) \cite{bulatov2022recurrent} to better handle ultra-long DNA sequences. By incorporating memory tokens and processing sequences in segments, RMT allows models to process sequences up to 196 kbp. \\

\noindent \textbf{Decoder-based approaches.} HyenaDNA~\cite{nguyen2024hyenadna} is a model capable of processing DNA sequences up to 1 million tokens at single nucleotide resolution. It leverages implicit convolutions to capture long-range dependencies, allowing for much longer sequence lengths than traditional dense attention. Evo~\cite{nguyen2024sequence} is a genomic foundation model designed to predict and generate DNA sequences at scales ranging from individual molecules to entire genomes. Utilizing the StripedHyena~\cite{stripedhyena} architecture, Evo incorporates both attention and data-controlled convolutional layers to process long DNA sequences efficiently, achieving a context length of 131 kbp. Trained on a massive dataset of prokaryotic genomes, Evo demonstrates competitive performance in zero-shot prediction tasks across DNA, RNA, and protein modalities. Caduceus~\cite{schiff2024caduceus} builds on the long-range Mamba block~\cite{gu2023mamba}, extending it to BiMamba for bi-directionality and MambaDNA for reverse-complement (RC) equivariance. In another work, by adapting large language models to genomic data, GenSLMs (Genome-Scale Language Models)~\cite{zvyagin2023genslms} can learn the evolutionary landscape of viral genomes. Pretrained on over 110 million prokaryotic gene sequences and fine-tuned with 1.5 million SARS-CoV-2 genomes, these models can accurately and rapidly identify variants of concern. GenSLMs utilize a hierarchical transformer-based model combining Generative Pretrained Transformers (GPT) and stable diffusion to capture both local and global genomic interactions. This approach enables effective whole-genome surveillance and the clustering of Variants of Concerns (VoC), aiding in timely public health interventions and vaccine development. For a more comprehensive review, we refer the reader to \cite{consens2023transformers} and \cite{benegas2024genomic}.\\

We note that the tokenization strategy and context length significantly influence the performance and applicability of the GFMs. High resolution tokenization allows these models to process DNA sequences at the single-nucleotide level, capturing fine-grained details essential for accurate predictions. Large context length is equally important as it enables the models to capture long-range dependencies within the DNA sequences. Specifically, genomic sequences often contain regulatory elements that can influence gene expression from a distance. These capabilities are vital for understanding complex genomic interactions and predicting the impact of genetic variants. Currently, state-space models like HyenaDNA~\cite{nguyen2024hyenadna} and Caduceus~\cite{schiff2024caduceus} are the primary solutions for handling ultra-long sequences at single nucleotide resolution. However, the potential of attention-based models in this domain remains an area worthy of exploration. 

\section{Methodology}
\label{sec:Methodology}
In this work, we present Gene42, an autoregressive decoder-only model that builds upon the LLaMA architecture \cite{touvron2023llama}. Consistent with the LLaMA design, our model features multiple layers, each incorporating multi-head attention mechanisms and feedforward neural networks. We maintain key components that contribute to LLaMA's effectiveness, including Rotary Position Embedding (RoPE) \cite{su2024roformer} for enhanced positional encoding, SwiGLU~\cite{shazeer2020glu} activation function for non-linear transformations, and RMSNorm pre-normalization \cite{touvron2023llama} for training stability, as detailed in Section \ref{sec:Architecture} and Table \ref{table:gene42-models-training-hyperparams}. A key contribution of Gene42 lies in its extended context window while maintaining a dense attention mechanism. While state-of-the-art dense attention models like NT \cite{dalla2023nucleotide} and GENA-LM \cite{fishman2023gena} process sequences up to 6 kbp and 36 kbp respectively, Gene42 pushes this limit to 192 kbp. We achieve this through an incremental continuous pretraining approach, starting with 4 kbp base models and progressively expanding their context window. Additionally, we adopt a character tokenization strategy similar to HyenaDNA \cite{nguyen2024hyenadna}. Gene42 models were trained on sequential DNA data derived and processed from the Human Reference Genome (GRCh38 genome assembly) and tokenized using a character tokenizer, as detailed in the following section.

\subsection{Data Preparation}

\subsubsection{Pretraining Data}
For training the Gene42 family of base models, we utilized the GRCh38\footnote{\url{https://www.ncbi.nlm.nih.gov/datasets/genome/GCF_000001405.26/}} human genome assembly, a comprehensive and well-annotated reference sequence. This assembly provides a robust and accurate representation of the human genome, making it an ideal foundation for training advanced genomic models. To ensure model robustness and effective performance evaluation, we split the data into training and evaluation sets. We allocated 99\% of the data for training and reserved 1\% for evaluation. The training dataset consists of ~3.5M (Million) sequences (each of 4,096 tokens), encompassing a total of 14.5B (Billion) nucleotides or tokens. This large volume of data exposes the model to a wide variety of genomic contexts and patterns, enhancing its ability to generalize across different genomic tasks. The evaluation dataset, though smaller, provides a significant sample for model assessment. It consists of 8,208 sequences with a total of ~33.6M tokens. This dataset is crucial for evaluating the model's next-token prediction accuracy, i.e., measuring the perplexity and reconstruction accuracy, as we will discuss in Section \ref{sec:perplexity}. To investigate the effects of dataset diversity, we incorporated genomic data from species other than human, previously used to train the GENA-LM model\footnote{\url{https://github.com/AIRI-Institute/GENA_LM/blob/main/data/full_ensembl_genomes_metadata.cvs}}. This multi-species dataset adds complexity and variety, aiding the model's ability to generalize across different genomic contexts. We processed a total of 8.1B tokens from 14 vertebrate species. The details of token distribution per species are provided in the Appendix in Table~\ref{table:ms-dataset}. The inclusion of multi-species data ensures that the model is not only adept at understanding human genomic sequences but also capable of identifying patterns and features conserved across different organisms. In summary, we used a combination of the GRCh38 and multi-species datasets to train our mixed dataset model, referred to as HRG+S in Table \ref{tab:GenomicBenchmarks} and Table \ref{tab:NT_Benchmark}.

\subsubsection{Sequence Tokenization and Data Preprocessing}

Sequence data tokenization at the nucleotide level using a Character tokenizer involves converting each nucleotide in a DNA sequence — Adenine (A), Cytosine (C), Guanine (G), and Thymine (T) — into individual tokens. This fine-grained approach allows models to capture the precise sequence of nucleotides, maintaining the highest possible resolution of the genetic information. Character tokenization is straightforward and effective for genomic data because the primary vocabulary consists of only a few characters (A, C, G, T), simplifying the tokenization process and ensuring no loss of information. By preserving the exact order and composition of nucleotide, Character tokenization enables models to detect subtle patterns, mutations, and regulatory elements within the DNA \cite{nguyen2024hyenadna}. This level of resolution or detail is crucial for tasks such as predicting the binding sites of transcription factors, identifying splice sites, and detecting genetic variants.

\subsection{Model Architecture and Pretraining}
\label{sec:Architecture}

The Gene42 models utilize a LLaMA-style architecture, specifically an autoregressive transformer decoder model \cite{touvron2023LLaMA2openfoundation}. All Gene42 base models are pretrained with a context length of 4,096 tokens. We refer to the 500M parameter model as \textbf{Gene42-B} (base size), and the 1.1B parameter models as \textbf{Gene42-L} (large). The Gene42-L models are further differentiated based on the context length achieved through incremental continuous pretraining: Gene42-L-4k (base model), Gene42-L-32k, Gene42-L-65k, and Gene42-L-192k. Table \ref{table:gene42-models-training-hyperparams} outlines the high-level model architecture and the comprehensive set of pretraining hyperparameters used to train the Gene42 family of models. 

\subsubsection{Base Models Pretraining on Cerebras CS-2 System}
We utilize Cerebras CS-2 for our training runs. The Cerebras CS-2 system is an AI accelerator that features 850,000 AI optimized compute cores, 40GB of on-chip SRAM, 20 PB/s memory bandwidth, and 220 PB/s interconnect \cite{cs2specs}. Motivated by advancements in natural language processing that suggest an increase in model parameter count correlates with improved performance~\cite{kaplan2020scaling}, we trained models with 500M and 1.1B parameters.

For the pretraining runs, we employ the AdamW optimizer with parameters $\beta_1 = 0.9$, $\beta_2 = 0.95$, and $\epsilon = 1 \times 10^{-8}$ \cite{loshchilov2019decoupledweightdecayregularization}. Additionally, we apply a weight decay of 0.1 across all Gene42 models. The learning rate is managed using a linear schedule for the warm-up phase, followed by a cosine decay schedule. The complete configuration for the pretraining phase is detailed in Table \ref{table:gene42-models-training-hyperparams}. 
\begin{table}[ht!]
\centering
\scriptsize
\addtolength{\tabcolsep}{0pt}
\def\arraystretch{1.2}
    \begin{tabular}{l| c  c |  c c c c}

            \textbf{Model} &  \textbf{Gene42-B} &  \textbf{Gene42-L-4k} &  \textbf{Gene42-L-32k} &  \textbf{Gene42-L-65k} &  \textbf{Gene42-L-192k} \\
            \shline
            \# of parameters & 500M & 1.1B & 1.1B & 1.1B & 1.1B \\
            Sequence Length & 4,096 bp & 4,096 bp & 32,768 bp & 65,356 bp & 192,000 bp \\
            Batch size & 128 & 128 & 32 & 32 & 32 \\
            Base Frequency & 10k & 10k & 15m & 15m & 15m \\
            Hidden size & 1,408 & 2,048 & 2,048 & 2,048 & 2,048 \\
            \# of hidden layers & 16 & 24 & 24 & 24 & 24 \\
            \# of attention heads & 16 & 32 & 32 & 32 & 32 \\
            Transformer FFN Dim. & 5632 & 5440 & 5440 & 5440 & 5440 \\
        \hline
            Optimizer & AdamW & AdamW & AdamW & AdamW & AdamW \\
            Betas & 0.9, 0.95 & 0.9, 0.95 & 0.9, 0.95 & 0.9, 0.95 & 0.9, 0.95\\
            Eps & 1e-8 & 1e-8 & 1e-8 & 1e-8 & 1e-8\\
            Weight Decay & 0.1 & 0.1 & 0.1 & 0.1 & 0.1\\
            Max grad norm & 1 & 1 & 1 & 1 & 1\\
            Learning rate (Linear) & 0 to 4.8e-4 & 0 to 4.8e-4 & 0 to 1e-4 & 0 to 1e-4 & 0 to 1e-4\\
            Iterations (Linear) & 0 to 1000 & 0 to 1000 & 0 to 50 & 0 to 20 & 0 to 20 \\
            Learning rate (Cosine) & 4.8e-4 to 4.8e-5 & 4.8e-4 to 4.8e-5 & 1e-4 to 4e-5 & 1e-4 to 4e-5 & 1e-4 to 4e-5 \\
            Iterations (Cosine) & 1000 to 19000 & 1000 to 19000 & 50 to 1000 & 20 to 200 & 20 to 200 \\
        
    \end{tabular}
    \caption{\textbf{Hyperparameters used for pretraining (left) and continuous pretraining (right)} of the Gene42 models.}
    \label{table:gene42-models-training-hyperparams}
\end{table}

\subsubsection{Continuous Pretraining for a Larger Context Length}
Training with extended context lengths incurs significant computational overhead, primarily due to the quadratic complexity of attention mechanisms. Recent research indicates that long context capabilities can be acquired through continuous pretraining, starting from models initially trained with shorter context lengths \cite{xiong-etal-2024-effective}. In our work, we employ continuous pretraining to extend our model's context length by reducing the rotation angle and base frequency. A smaller rotation angle mitigates the decaying influence of RoPE on distant tokens, thereby enhancing the model's ability to handle longer contexts effectively.

We continually pretrain the Gene42-L model by starting from our checkpoint with a context length of 4,096 tokens and then progressively increasing it to 8,192 tokens, followed by 16,384, 32,768, 65,536, 131,072, and finally reaching 192,000 tokens. We apply a linear warm-up schedule followed by a cosine decay schedule for our continuous pretraining runs. Detailed hyper-parameter configurations for continuous pretraining of the 32k, 65k, and 192k context length models can be found in the right part of Table \ref{table:gene42-models-training-hyperparams}.

\subsection{Model Evaluation Using Perplexity}
\label{sec:perplexity}

A crucial step in our model development process is the evaluation of Gene42 models' performance before validating them on downstream tasks. To achieve this, we employ the Perplexity (or PPL) metric, a widely used measure for assessing autoregressive or causal language models. Perplexity provides insight into how well a model predicts the next token. For a tokenized sequence $X=\{x_0,x_1,\dots,x_n \}$, Perplexity is defined as the exponentiated average negative log-likelihood of a sequence: $\text{PPL}(X) = \exp \{- \frac{1}{t} \sum^{t}_{i} \log p_{\theta} (x_{i}|x_{<i}) \}$, where $\log p_{\theta}$ is the log-likelihood of the i$^{th}$ token conditioned on the preceding tokens $x_{<i}$ according to our model, and $t$ is the context length. 

\begin{figure}[ht!]
    \centering
    \includegraphics[width=\linewidth]{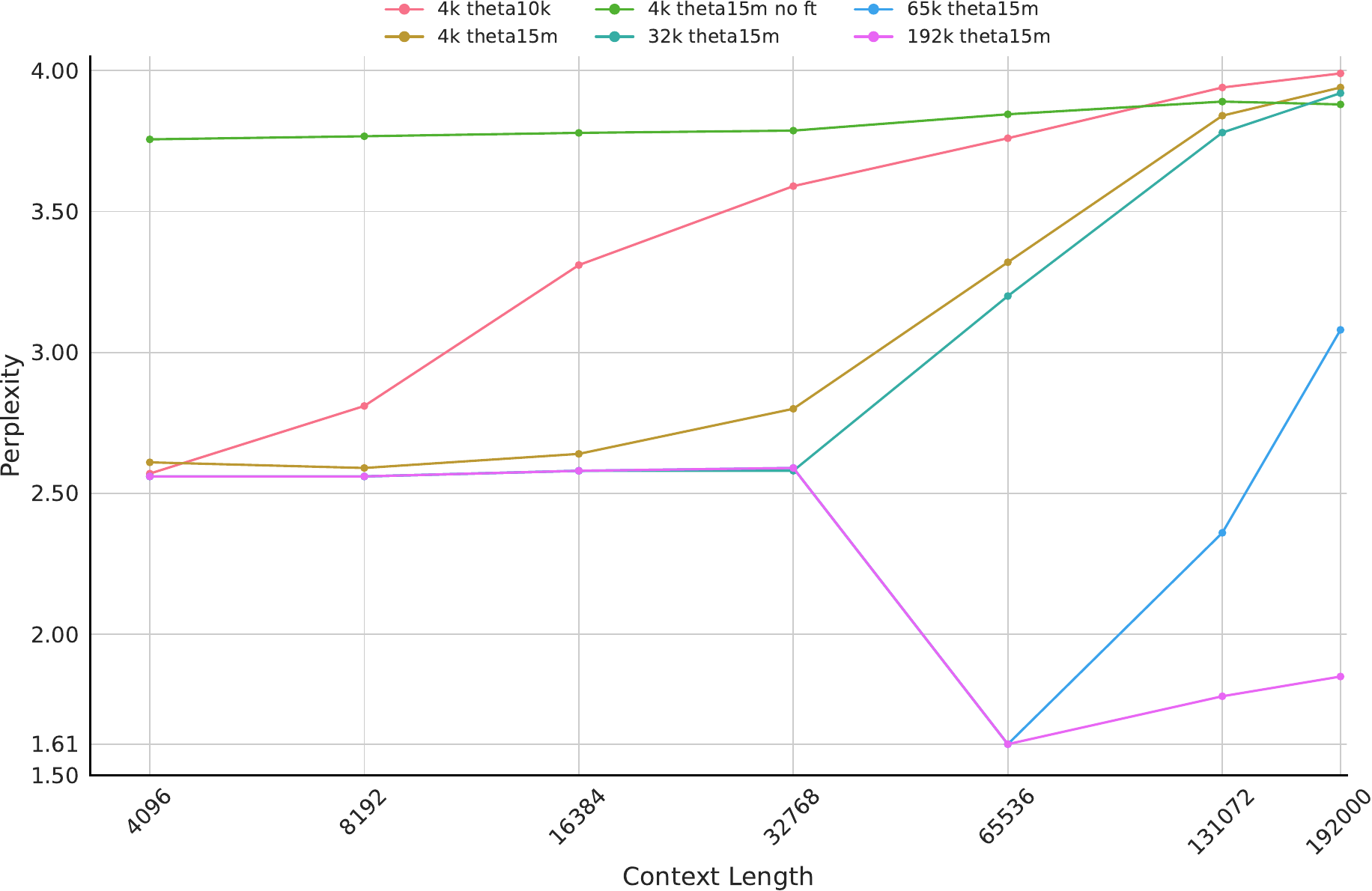} 
    \caption{\textbf{Perplexity of Gene42-L models on the evaluation dataset}. The input sequence lengths are varied from 4k to 192k. Shorter context models (4k-32k) have very high perplexity for long sequences. Longer context models (65k-192k) maintain lower perplexity, with the 65k and 192k models achieving optimal performance of 1.61 at 65k context length.}
    \label{fig:perplexity}
\end{figure}

Figure~\ref{fig:perplexity} summarizes the PPL values computed for all Gene42-L models, with context lengths ranging from 4k to 192k, on 1\% of the HRG evaluation data. First, we can see the limited capabilities of short-range context length models to model long-range sequences, attested by the increase in perplexity values as the sequence length increases (red, yellow, and green lines). Second, Gene42-L long-range context length models (65k and 192k) built by continuously pretraining the base model trained on 4 kbp sequences achieved an optimal perplexity of 1.61, which corresponds to a reconstruction accuracy equal to 0.789, on sequences of length 65 kbp. Third, the 192k context length model achieved slightly higher perplexity values of 1.78 (reconstruction accuracy of 0.723) and 1.85 (reconstruction accuracy of 0.70) on longer sequence lengths of 131k and 192k, respectively. This suggests that a model trained with a sequence length of 65 kbp had the best performance measured in the next base pair prediction in the training data. We believe it is possible to achieve lower perplexity values on the longer sequence lengths by increasing the model size further.

These results provide valuable insights when compared to other GFMs. For instance, the work on HyenaDNA \cite{nguyen2024hyenadna} reported significantly higher perplexity values, ranging between 2.9 and 3, for long sequences in the range of 100k and 1 million tokens. This contrast highlights the efficiency of our Gene42 models in capturing long-range dependencies in genomic sequences. On the other hand, the GenSLMs work \cite{zvyagin2023genslms}, which focuses on prokaryotic genes (viruses and bacteria), reported lower perplexity values close to 1 on their validation dataset. This difference can be attributed to the inherent characteristics of prokaryotic gene sequences, which are considerably shorter than human genomic sequences. The simpler structure of prokaryotic genomes likely contributes to the lower perplexity values observed in that study. In comparison, our results suggest that Gene42 effectively balances the complexity of human genomic sequences with the ability to model long-range dependencies.            

\section{Experimental Results}
\label{sec:experiments}

This section presents a comprehensive evaluation of our Gene42 models across a diverse array of genomic downstream tasks. Our evaluation strategy is twofold: first, we assess the quality of embeddings produced by our pretrained models through biotype classification, which provides insight into the model's ability to capture fundamental genomic features acquired during pretraining. Second, we evaluate the models' adaptability to specific genomic tasks through finetuning experiments.

We systematically compare the performance of Gene42 models against state-of-the-art models, including NT~\cite{dalla2023nucleotide}, HyenaDNA~\cite{nguyen2024hyenadna}, Caduceus~\cite{schiff2024caduceus}, DNABERT v2~\cite{zhou2023dnabert}, and Gena-LM~\cite{fishman2023gena}. This comparative analysis spans both short-range and long-range genomic experiments, providing a holistic view of our models' capabilities across various sequence lengths and task categories. These experiments aim to demonstrate the effectiveness of Gene42 models in capturing genomic information and their adaptability to specific downstream tasks.   

\subsection{Biotype Classification}

In this section, we present the experimental results for biotype sequence classification. Embeddings were extracted from the Gene42-B, Gene42-L-32k, and Gene42-L-65k models for these sequences. These embeddings were then used as input features for an XGBoost classifier~\cite{chen2016xgboost}. The classification performance was evaluated using the F1 score, comparing results across the models.

Biotype datasets, such as those from the Ensembl database~\cite{cunningham2022ensembl}, provide comprehensive annotations of gene and transcript types based on experimental evidence and computational predictions. These biotypes include classifications such as protein-coding genes, long noncoding RNAs, pseudogenes, and small non-coding RNAs, each playing distinct roles in biological processes. Accurately differentiating between these biotypes through distinct embeddings demonstrates the model's ability to capture semantic relationships and functional properties of genes. This capability is crucial for understanding complex biological systems and improving the interpretation of genomic data. Biotype embeddings were generated using data from human genes annotated with various biological functions from Ensembl. To create a balanced dataset, we capped all the classes at a maximum of three thousand samples. The final embedding was generated using max pooling over the token dimension. When a sequence exceeded the model length, it was chunked, embeddings were produced for each chunk, and these embeddings were then averaged across chunks.

\begin{table}[ht!]
\centering
\footnotesize
\addtolength{\tabcolsep}{0pt}
\def\arraystretch{1.2}
\begin{tabular}{l  c  c  c}
\textbf{Model} &  \textbf{Max Seq Len (bp)} & \textbf{F1 score} \\
\shline
DNABERT v2 &  512 & 0.654 \\
NT 50M   & 12,282 & 0.681 \\
NT 2.5B & 5,994 & 0.759 \\
HyenaDNA-medium & 160,000 & 0.709 \\
\hline

 \textbf{Gene42-B} & 4,096 & 
\bf 0.763 \\
 \textbf{Gene42-L-32k}  & 32,678 & \bf 0.771 \\
 \textbf{Gene42-L-65k} &  65,356 & \bf 0.782 \\
\end{tabular}
\caption{\textbf{Biotype classification results.} Gene42 models show superior results to other competitors. Increasing the context length improves the performance of Gene42 models.}
\label{tab:f1-score}
\end{table}

Gene42-L models achieved notable F1 scores of 0.771 and 0.782 for context lengths of 32k and 65k, respectively. These results highlight the effectiveness of Gene42 models in capturing essential biological information from sequence data, outperforming DNABERT v2, two variants of Nucleotide Transformer with 50M and 2.5B parameters, and HyenaDNA. The highest F1 score of 0.782, achieved with a context length of 65,356 bp, demonstrates the proficiency of the Gene42 model in representing biological features, particularly with longer context lengths. 

\subsection{Finetuning Experiments}

To assess the performance of our Gene42 genomic foundation models, we conducted evaluations on a series of downstream genomic applications. These benchmarks include critical tasks such as promoter prediction, chromatin profiling, splice site detection, and enhancer activity prediction. Each task evaluates the model's ability to interpret and predict functional elements within DNA sequences, which are essential for understanding gene regulation and genomic function.

\subsubsection{Genomic Benchmarks}

We started our evaluation with short-range genomics data using the Genomic Benchmarks \cite{grevsova2023genomic}. This benchmark suite comprises eight regulatory element classification datasets, including promoters, enhancers, and open chromatin regions, derived from three model organisms: human, mouse, and roundworm. The sequence lengths in these datasets range from 200 to 500 bp, with one dataset extending up to 4,776 bp. We assessed various Gene42 models and compared their performance to the HyenaDNA and Caduceus models. The evaluation metric used was the Top-1 accuracy score (in \%).

\begin{table}[ht!]
\centering
\scriptsize
\addtolength{\tabcolsep}{0pt}
\def\arraystretch{1.2}
\begin{tabular}{l|c|c|c|c|c|c|c} 
\textbf{DATASET} & HyenaDNA & Caduceus & \multicolumn{5}{c}{\textbf{Gene42}} \\
\hline
Training data & \multicolumn{3}{c|}{\textbf{HRG}} & HRG+S & \multicolumn{3}{c}{HRG} \\
\hline
\# of parameters &  1.6M & 1.9M & \multicolumn{3}{c|}{\textbf{500M}} & \multicolumn{2}{c}{1.1B}\\
\hline
Context length (bp) &  1,024 &  131,000 & \multicolumn{2}{c|}{\textbf{4,096}} & 192,000 & 32,768 & 192,000 \\
\shline
Mouse enhancers & \textbf{85.1} & 75.4 & 79.4 & \cb 82.4 & 77.8 & 73.9 & 73.9 \\
Coding vs. Intergenomic & 91.3 & 91.5  & \textbf{95.3} & \textbf{95.2} & \textbf{94.9} & \bf \cb 95.6 & \textbf{94.3} \\
Human vs. Worm & 96.6 & 97.3  & \cb \textbf{97.5} & 97.3 & 97.3 & \cb \textbf{97.5} & 96.6 \\
Human Enhancers Cohn & 74.2 & \textbf{74.7} & \cb 74.4 & 73.8 & 73.3 & 74.1 & 71.6 \\
Human Enhancers Ensembl & 89.2 & 89.3 & \textbf{91.6} & \cb \textbf{91.9} & \textbf{91.4} & \textbf{91.7} & \textbf{90.1} \\
Human Regulatory & 93.8 & 87.2 & \textbf{94.5} & \cb \textbf{96.3} & \textbf{94.5} & \textbf{94.4} & \textbf{94.2} \\
Human non-TATA Promoters & 96.6 & 94.6 & \textbf{97.6} & 96.3 & \textbf{96.9} & \cb \textbf{97.7} & 94.1 \\
Human OCR Ensembel & 80.9 & \textbf{82.8} & \cb 80.9 & \cb 80.9 & 80.0 & 79.8 & 76.4 \\
\shline
\textbf{Average accuracy (\%)} & 88.5 & 86.5 & \textbf{88.9} & \cb \textbf{89.3} & 88.3 & 88.1 & 86.4 \\

\end{tabular}
\caption{\textbf{Evaluation on Genomic Benchmarks} \cite{grevsova2023genomic} - Top-1 accuracy (\%) of Gene42 models compared to HyenaDNA and Caduceus. SoTA performances are taken from \cite{nguyen2024hyenadna} and \cite{schiff2024caduceus}, respectively. \textbf{Bold}: best performance; \colorbox{green!30}{Green} indicates best performance within the Gene42 models.}
\label{tab:GenomicBenchmarks}
\end{table}

It has been demonstrated in \cite{nguyen2024hyenadna} that HyenaDNA outperforms CNN, GPT,  DNABERT language models on most tasks. Our results show that Gene42, trained on both Human Reference Genome (HRG) and multi-species data, achieves a top-1 accuracy score of 95.3\% on this task, surpassing both DNABERT (92.5\%) and HyenaDNA (91.3\%). Consequently, Gene42-B/L models set a new state-of-the-art performance on 5 datasets over 8 on the Genomic Benchmarks.   

\subsubsection{Nucleotide Transformer Benchmark}

To further evaluate Gene42 models with different numbers of parameters (500M and 1.1B) and different context lengths (4 kbp, 32 kbp, and 192 kbp), we fine-tuned them on downstream tasks defined by the NT Benchmark and compared results to other models, i.e., NT model (2.5B trained on Multispecies data) \cite{dalla2023nucleotide} and HyenaDNA model \cite{nguyen2024hyenadna} (1.6M parameters model with 1 kbp and 32 kbp context lengths). To allow fair comparison, we consider results as they are reported in the NT Leaderboard\footnote{\url{https://huggingface.co/spaces/InstaDeepAI/nucleotide_transformer_benchmark}}. A summary of the results is reported in Table~\ref{tab:NT_Benchmark}.

\begin{table}[ht!]
\centering
\scriptsize
\addtolength{\tabcolsep}{0pt}
\def\arraystretch{1.2}
\begin{tabular}{l| c | c | c | c | c | c | c | c | c | c }

\textbf{DATASET} & \multicolumn{3}{c|}{Nucleotide Transformer (NT)}  & \multicolumn{2}{c|}{HyenaDNA} & \multicolumn{5}{c}{\textbf{Gene42}}\\
\hline
Training data & HRG & 1000G & MS & \multicolumn{3}{c|}{\textbf{HRG}} & HRG+S & \multicolumn{3}{c}{\textbf{HRG}}\\
\hline
\# of parameters & 500M & \multicolumn{2}{c|}{2.5B} & \multicolumn{2}{c|}{1.6M}& \multicolumn{3}{c|}{\textbf{500M}} & \multicolumn{2}{c}{\textbf{1.1B}} \\
\hline
 Context length & \multicolumn{3}{c|}{6 kbp} & 1 kbp & 32 kbp & \multicolumn{2}{c}{\textbf{4 kbp}} & 192 kbp & 32 kbp & 192 kbp\\
\shline
H3 & 0.718 & 0.755 & 0.791 & 0.779 & 0.741 & 0.791 & 0.790 & 0.778 & \cb \bf 0.800 & 0.732 \\

H3K4me1 & 0.377 & 0.418 & \textbf{0.544} & 0.512 & 0.383 & \cb 0.532 & 0.508 & 0.481 & 0.509 & 0.371 \\

H3K4me2 & 0.262 & 0.279 & 0.322 & 0.455 & 0.275 & \cb \bf 0.472 & 0.327 & 0.445 & 0.454 & 0.269 \\

H3K4me3 & 0.237 & 0.309 & 0.408 & \textbf{0.549} & 0.290 & \cb 0.516 & 0.439 & 0.427 & 0.32 & 0.185 \\

H3K9ac & 0.476 & 0.494 & 0.550 & 0.581 & 0.470 & 0.605 & 0.567& \cb \bf 0.614 & 0.558& 0.478 \\

H3K14ac & 0.382 & 0.452 & 0.537 & 0.612 & 0.404 & 0.609 & 0.564 & 0.626 & \cb \bf 0.635 & 0.355 \\

H3K36me3 & 0.465 & 0.526 & 0.616 & 0.613 & 0.476 & 0.644 & 0.611 & 0.632 & \cb \bf 0.673 & 0.509 \\

H3K79me3 & 0.560 & 0.574 & 0.621 & 0.672 & 0.570 & \cb \bf 0.706 & 0.668 & 0.687 & \cb \bf 0.706 & 0.618 \\

H4 & 0.748 & 0.788 & \textbf{0.807} & 0.763 & 0.760 & 0.800 & \cb 0.803 & 0.771 & 0.799 & 0.773 \\

H4ac & 0.382 & 0.452 & 0.537 & \textbf{0.612} & 0.404 & \cb 0.593 & 0.535 & 0.591 & 0.561 & 0.365 \\

\textbf{Avg. (MCC)} & 0.461 & 0.505 & 0.573 & 0.615 & 0.477 & \cb \bf 0.627 & 0.581 & 0.605 & 0.602 & 0.465 \\

\shline
Prom. all & 0.951 & 0.965 & \textbf{0.975} & 0.960 & 0.956 & 0.969 & \cb 0.971 & 0.966 & 0.969 & 0.963 \\

Prom. non-TATA & 0.950 & 0.967 & \textbf{0.976} & 0.959 & 0.954 & 0.970 & \cb 0.972 & 0.967 & 0.969 & 0.963 \\

Prom. TATA & 0.937 & 0.955 & 0.959 & 0.944 & 0.939 & \cb \bf 0.964 & 0.963 & 0.951 & 0.954 & \cb \bf 0.964 \\

\textbf{Avg. (F1-score)} & 0.946 & 0.962 & \textbf{0.970} & 0.954 & 0.950 & \cb 0.968 & \cb 0.968 & 0.961 & 0.964 & 0.963 \\
\shline
Enhancer & 0.501 & 0.542 & 0.548 & 0.517 & 0.494 & 0.543& 0.517 & 0.523 & \cb \bf 0.557 & 0.47 \\

Enhancer types & 0.429 & 0.435 & 0.450 & 0.386 & 0.358 & \cb \bf 0.463 & 0.434 & 0.443 & 0.461 & 0.413 \\

\textbf{Avg. (MCC)} & 0.465 & 0.489 & 0.499 & 0.452 & 0.426 & 0.503 & 0.475 & 0.483 & \cb 0.509 & 0.441 \\
\shline
Splice acceptor & 0.962 & 0.965 & \textbf{0.987} & 0.958 & 0.958 & 0.977 & \cb 0.983 & 0.973 & 0.981 & 0.975 \\
Splice donor & 0.970 & 0.979 & \textbf{0.987} & 0.949 & 0.958 & 0.979 & \cb 0.985 & 0.979 & 0.982 & 0.981 \\
Splice all & 0.971 & 0.969 & \textbf{0.982} & 0.956 & 0.962 & \cb 0.981 & \cb 0.981 & 0.978 & \cb 0.981 & 0.977 \\
\textbf{Avg. (F1-score)} & 0.968 & 0.971 & \textbf{0.985} & 0.954 & 0.959 & 0.979 & \cb 0.983 & 0.976 & 0.981 & 0.977 \\
\end{tabular}
\caption{\textbf{Results on Nucleotide Transformer Benchmark}~\cite{dalla2023nucleotide}. The Matthews correlation coefficient (MCC) is used as the performance metric for the enhancer and epigenetic marks dataset, and the F1-score is used for the promoter and splice site dataset. \textbf{Bold}: best performance; \colorbox{green!30}{Green} indicates best performance within the Gene42 models.}
\label{tab:NT_Benchmark}
\end{table}
HyenaDNA (1k) surpasses others on 2 datasets out of 18, Multispecies NT (2.5B parameters model with 6k context length) outperforms on 8 datasets, with the highest score on the Splice datasets. The Gene42-B model with 500M parameters and 4 kbp context length, trained solely on human data, also achieves the highest scores on 8 datasets.

Despite their architectural differences (BERT-based for NT and LLaMA-based for Gene42), both models employ dense self-attention, enabling them to capture global contextual information effectively. This mechanism proves highly effective for modeling complex dependencies within sequences but comes with significant computational and memory overhead, particularly for long sequences, due to its quadratic complexity. In contrast, HyenaDNA leverages the Hyena operator, which utilizes implicit convolutions to capture dependencies more efficiently, reducing computational and memory requirements. While dense self-attention achieves superior results on short-range genomic tasks (as evidenced in Table~\ref{tab:NT_Benchmark}), the Hyena operator in HyenaDNA offers a more efficient alternative, though it comes with some performance trade-offs.       

\subsubsection{Variant Pathogenicity Classification}

The task of variant pathogenicity prediction assesses the likelihood that genetic variants contribute to disease. This task is essential for understanding the molecular basis of genetic disorders and guiding the development of targeted therapies. The variant pathogenicity classification broadly results in a variant being classified as benign (including likely benign), pathogenic (including likely pathogenic), or as a variant of uncertain significance (VUS). We trained our Gene42 models (500M and 1.1B parameter variants) for pathogenicity classification using the ClinVar dataset \cite{landrum2014clinvar}, following training protocols similar to those in \cite{sayeed2024gene}. This dataset, derived from the Hg38 human genome assembly, is often considered the primary source for variant annotation. Complete finetuning was performed on a training dataset comprising 64.5K samples and a test dataset of 8.8K samples, each with a 1K sequence length. We employed a learning rate of 1e-4 with a linear scheduler for complete finetuning.

\begin{table}[ht!]
\centering
\scriptsize
\addtolength{\tabcolsep}{0pt}
\def\arraystretch{1.2}
\begin{tabular}{l | c c | c c | c | c c c c c}
\textbf{Model} & \multicolumn{2}{c|}{NT$^\dag$} & \multicolumn{2}{c|}{GENA-LM$^\dag$} & HyenaDNA$^\dag$ & \multicolumn{5}{c}{\textbf{Gene42}} \\
\hline
Experiment & \multicolumn{5}{c|}{FT (finetuning)} & FT & ZSL & \textbf{FT(1k)}  & FT(2k) & \textbf{RF}\\
\hline
\# of Parameters & 500M & 100M & lastln & Large & 1.6M & 500M & \multicolumn{4}{c}{\textbf{1.1B} }\\
\shline
Accuracy (\%) & 78.2 & \textbf{90.6} & 67.8 & -- & 56.6 & 67.9 & 38.9 & 81.4 & \cb 85.2 & 78.0 \\
Precision (\%) & 82.1 & 91.1 & \textbf{94.8} & -- & 66.5 & 50.3 & 32.6 & 65.0 & 73.0 & \cb 82.6 \\
Recall (\%) & 80.1 & 90.8 & 79.0 & -- & 43.3 & \cb \textbf{92.9} & 82.8 &  92.4  & 86.5 & 78.0 \\
\hline
AUC ROC  & -- & 0.892 & -- & 0.731 & 0.608 & 0.840 & 0.513 & \cb \textbf{0.931} & 0.925 & 0.915 \\
AUC PR & -- & \textbf{0.851} & -- & 0.717 & 0.477 & 0.679 & 0.605 & 0.799 & \cb 0.819 & -- \\
\end{tabular}
\caption{\textbf{Evaluations on variant pathogenicity classification task}~\cite{sayeed2024gene} and comparative performances of various models based on Accuracy, Precision/Recall, AUC ROC and AUC PR. The last column reports results of Random Forest (RF) trained on Gene42 Embedding. \textbf{Note:} (1k) and (2k) refer to the evaluation at 1000 and 2000 steps, respectively; ZSL: Zero-shot learning. $^\dag$Performances reported previously in \cite{sayeed2024gene}. \textbf{Bold} indicates best performance. \colorbox{green!30}{Green} indicates best performance within the Gene42 models.}
\label{tab:gene_pt}
\end{table}

We compared the performance of models using average accuracy, precision, recall, AUC-ROC, and AUC-PR as evaluation metrics. Table~\ref{tab:gene_pt} presents a comparative analysis of our Gene42 models against HyenaDNA \cite{nguyen2024hyenadna}, Nucleotide Transformer \cite{dalla2023nucleotide}, and GenaLM \cite{fishman2023gena} across finetuning (FT), zero-shot (ZS), and embedding (Emb) scenarios. The results for the Random Forest model trained on Gene42 embeddings are also included. Our 500M parameter model achieved the highest recall, while the 1.1B parameter variant showed robust performance at the 2000th-step and the highest AUC-ROC value of 0.931 when evaluated after 1000 steps. 

\subsubsection{Chromatin Profiling}

The chromatin profiling challenge encompasses understanding the complex organization and regulatory mechanisms of chromatin, which involves DNA and protein interactions within the cell nucleus. This challenge is multifaceted, involving the characterization of chromatin states, histone modifications, DNA methylation, and chromatin accessibility. The chromatin profiling prediction problem involves developing models that accurately predict the regulatory states and functions of chromatin regions across the genome. This problem is crucial for understanding gene regulation, cellular differentiation, and the mechanisms underlying various diseases. For example, DeepSEA \cite{zhou2015deepsea} predicts genomic variant effects on a wide range of chromatin features at the variant position, including Transcription Factors (TF) binding, DNase I Hypersensitive Sites (DHS), and Histone Marks (HM), in multiple human cell types.

Here, we evaluate the Gene42 family of models on the chromatin profile prediction task. This is a multi-label classification task with 919 labels. Using the DeepSEA \cite{zhou2015deepsea} dataset, we fine-tuned our models on 4.4 million sequences of 1,000 and 8,000 bp and evaluated them on an unseen test dataset of 455,000 samples of the same lengths. Our models outperform the previous benchmarks (Table \ref{table:chromatin-profile-benchmark}) on the Transcription Factor binding profiles (TF) and DNase I-hypersensitive sites (DHS) tasks.

\begin{table}[ht!]
\centering
\scriptsize
\addtolength{\tabcolsep}{0pt}
\def\arraystretch{1.2}
    \begin{tabular}{l| c | c | c | c | c | c |c}
            \textbf{Task} & DeepSEA$^\dag$ & HyenaDNA$^\dag$ & \multicolumn{2}{c|}{ \textbf{Gene42}}& Bigbird$^\dag$ & HyenaDNA$^\dag$ & \textbf{Gene42} \\
\shline       
Sequence length  & \multicolumn{4}{c|}{\textbf{1,000 bp}} & \multicolumn{3}{c}{\textbf{8,000 bp}} \\
\hline
\# of Parameters & 40M & 7M & 500M & \textbf{1.1B}  & 110M & 3.5M & \textbf{1.1B}   \\
\hline
Context length & -- & 1,024 & 4,096 & \textbf{32,768} &  & 32,770 &  \textbf{32,768}\\
\shline
            \textbf{TF} (Median AUC) & 0.958 & 0.964 & 0.966 & \textbf{0.967}  & 0.961 & 0.955 & \textbf{0.965}  \\
            \textbf{DHS} (Median AUC) & 0.923 & 0.930 & 0.932 & \textbf{0.934}  & 0.921 & 0.917 & \textbf{0.931}  \\
            \textbf{HM} (Median AUC) & 0.856 & \textbf{0.863} & 0.838 & 0.839 & 0.887 & \textbf{0.893} & 0.836 \\
    \end{tabular}\\
    \caption{\textbf{Chromatin profiling prediction:} Median AUC scores for Transcription factor binding profiles (\textbf{TF}), DNase I-hypersensitive sites (\textbf{DHS}), Histone marks (\textbf{HM}) with comparisons with DeepSEA and HyenaDNA. $^\dag$Performances for DeepSEA, Bigbird and HyenaDNA approaches taken from \cite{nguyen2024hyenadna}. \textbf{Bold}: best performance.}
    \label{table:chromatin-profile-benchmark}
\end{table}

We performed additional experiments to study the effect of finetuning only the classification head compared to finetuning all layers of the model. We observed that finetuning all layers of the model leads to significantly higher scores compared to finetuning only the classification head (Table \ref{table:chromatin-profile-ablation}). 

\begin{table}[ht!]
\centering
\scriptsize
\addtolength{\tabcolsep}{0pt}
\def\arraystretch{1.2}
    \begin{tabular}{l| c | c | c | c |  c | c }
                Sequence Length & \multicolumn{4}{c|}{\textbf{1,000 bp}} & \multicolumn{2}{c}{\textbf{8,000 bp}}\\
\hline
                \# of Params & \multicolumn{2}{c|}{500M} & \multicolumn{2}{c|}{1.1B} & \multicolumn{2}{c}{1.1B}  \\
\hline
                Context length & \multicolumn{2}{c|}{4,096} & \multicolumn{2}{c|}{32,768} & \multicolumn{2}{c}{32,768}\\
\hline
                finetuning & \textbf{All Layers} & Classif. Head & \textbf{All Layers} & Classif. Head & \textbf{All Layers} & Classif. Head \\
\shline 
                \textbf{TF} & \textbf{96.6} & 72.5 & \textbf{96.7} & 73.1 & \textbf{96.5} & 73.0\\
                \textbf{DHS} & \textbf{93.2} & 67.5 & \textbf{93.4} & 68.3 & \textbf{93.1} & 68.3 \\
                \textbf{HM} & \textbf{83.8} & 73.7 & \textbf{83.9} & 73.9 & \textbf{83.6} & 74.0 \\
    \end{tabular}
    \caption{\textbf{Comparing finetuning strategies on Chromatin profiling.} Finetuning all layers yields better performance in all cases (in \textbf{bold}).}
    \label{table:chromatin-profile-ablation}
\end{table}

\subsubsection{Species Classification}

Species classification based on gene sequences involves analyzing genetic material to identify unique markers and patterns that differentiate species. This method relies on genomic regions that are highly conserved within a species but variable between different species. By leveraging the conserved nature of genomes and the advanced capabilities of our models, we aim to enhance the accuracy and efficiency of species classification based on genetic data. 

\begin{table}[ht!]
\centering
\footnotesize
\addtolength{\tabcolsep}{0pt}
\def\arraystretch{1.2}
\begin{tabular}{l| c  c}
\textbf{Model} & \textbf{Seq Len (bp)} & \textbf{Accuracy (\%)} \\
\shline
Transformer$^\dag$ (from \cite{nguyen2024hyenadna}) & 1,024 & 55.4 \\
HyenaDNA$^\dag$ \cite{nguyen2024hyenadna} & 1,024 & 61.1 \\
RMT+gena-lm-bert-base-t2t$^\dag$ \cite{kuratov2024recurrent} & 1,024 & 61.4 \\
\textbf{Gene42-L} & \textbf{1,024} & \textbf{66.9} \\
\shline
Transformer$^\dag$ (from \cite{nguyen2024hyenadna}) & 32,768 & 88.9 \\
HyenaDNA$^\dag$ \cite{nguyen2024hyenadna} & 32,768 & 93.4 \\
RMT+gena-lm-bert-base-t2t$^\dag$ \cite{kuratov2024recurrent} & 32,768 & 99.2 \\
\textbf{Gene42-L-32k} & \textbf{32,768} & \textbf{99.5} \\
\shline
Transformer$^\dag$ & 250,000 & \ding{55} \\
HyenaDNA$^\dag$ & 250,000 & 97.9 \\

\end{tabular}\\
\caption{\textbf{Species classification performance}. Note that the symbol \ding{55} represents infeasible training time. $^\dag$Performances reported previously in \cite{nguyen2024hyenadna} and \cite{kuratov2024recurrent}.}
\label{species cls}
\end{table}

For this five-way classification task, we added a linear head and performed a full finetuning of the model. We observed that the optimal learning rate across all settings was 1e-5, in conjunction with a cosine scheduler.
Following the experimental settings employed by HyenaDNA (\cite{nguyen2024hyenadna}), we randomly sampled DNA sequences from five species: human \textit{(Homo sapiens)},  mouse \textit{(Mus musculus)}, lemur \textit{(Lemur catta)}, pig \textit{(Sus scrofa)}, and hippo \textit{(Hippopotamus amphibius)} to predict the species label. We reserved four chromosomes (1, 3, 12, and 13) from each species for evaluation and utilized the remaining chromosomes for training. Our model, with a maximum context length of 4,096, fine-tuned on sequence length 1k to match their setting, outperformed HyenaDNA's results at the same sequence length. Additionally, when fine-tuned with a larger sequence length of 32,768, our model surpassed all results achieved by HyenaDNA \cite{nguyen2024hyenadna} and GENA-LM (RMT+gena-lm-bert-base-t2t) \cite{kuratov2024recurrent} across various context lengths of 32k, 250k, 450k, and 1 Million (see Table~\ref{species cls}).

\section{Conclusion}
\label{sec:Conclusion}

In this paper, we introduce Gene42, a novel family of high-resolution Genomic Foundation Models designed to handle short, medium, and long-range context lengths at a single-nucleotide resolution. By leveraging a LLaMA-style architecture and single-nucleotide tokenization, our models achieved significant improvements in capturing long-distance dependencies in DNA sequences. Through continuous pretraining, we extended the context length of our models to encompass up to 192,000 bases. Extensive evaluations demonstrated that Gene42 models outperform existing state-of-the-art models. The ability to process ultra-long genomic sequences while maintaining dense attention modules positions Gene42 as a superior tool for genomic analysis, with potential applications in precision medicine and genetic research. 

Exploring the application of Gene42 models in clinical research for personalized medicine or predicting patient-specific responses to therapies and identifying novel therapeutic targets, holds great potential for advancing healthcare.

\bibliographystyle{alpha}
\bibliography{main}

\newpage

\section*{Appendix}

\subsubsection*{Multi-species Dataset Distribution}
This section presents the distribution of species in our Multi-species dataset. The dataset encompasses 14 distinct vertebrate species, collectively contributing 8 billion tokens of genomic data.
\begin{table}[ht!]
    \centering
\scriptsize
\addtolength{\tabcolsep}{0pt}
\def\arraystretch{1.2}
    \begin{tabular}{l c}
                Species & Num. of Tokens \\
                \shline
                Acanthochromis-polyacanthus-ASM210954v1 & 614M \\
                Accipiter-nisus-Accipiter-nisus-ver1.0 & 561M \\
                Ailuropoda-melanoleuca-ASM200744v2 & 215M \\
                Amazona-collaria-ASM394721v1 & 1221M \\
                Amphilophus-citrinellus-Midas-v5 & 681M \\
                Amphiprion-ocellaris-AmpOce1.0 & 788M \\
                Amphiprion-percula-Nemo-v1 & 48M \\
                Anabas-testudineus-fAnaTes1-2 & 28M \\
                Anas-platyrhynchos-ASM874695v1 & 216M \\
                Anas-platyrhynchos-platyrhynchos-CAU-duck1-0 & 197M \\
                Anolis-carolinensis-AnoCar2-0v2 & 256M \\
                Anas-zonorhyncha-ASM222487v1 & 1134M \\
                Anser-brachyrhynchus-ASM259213v1 & 1093M \\
                Anser-cygnoides-GooseV1-0 & 1067M \\
    \end{tabular}
    \caption{Multi-species dataset.}
    \label{table:ms-dataset}
\end{table}

\subsubsection*{Hyperparameters for Downstream Tasks}

In this section we provide the hyper-parameter settings used for finetuning the Gene42 models on different downstream tasks. 

\begin{table}[ht!]
\centering
\scriptsize
\addtolength{\tabcolsep}{0pt}
\def\arraystretch{1.2}
    \begin{tabular}{l c c}
            & \multicolumn{2}{c}{ \textbf{Gene42}} \\  
        \shline
            \# of parameters & 500M & 1.1B \\
            Sequence Length & 4,096 & 32,768 \\
        \shline
            Optimizer & AdamW & AdamW \\
            Learning rate & 0.0001 & 0.0001 \\
            Betas & 0.9, 0.999 & 0.9, 0.999 \\
            Weight Decay & 0 & 0 \\
            Epochs & 2 & 2 \\
            Scheduler & Linear & Linear \\
            Warmup Ration & 0.1 & 0.1
        
    \end{tabular}
    \caption{Hyperparameters used for finetuning Gene42 models on Chromatin profiling prediction.}
    \label{table:chromatin-profile-finetune}
\end{table}

\begin{table}[ht!]
    \centering
    \scriptsize
    \addtolength{\tabcolsep}{0pt}
    \def\arraystretch{1.2}
    \begin{tabular}{l c c c c}
            \bf Hyperparameter & \bf Value \\
            \shline
            Optimizer & AdamW \\
            Learning rate & 1e-4 \\
            Betas & (0.9, 0.999)\\
            Weight Decay & 0 \\
            Epochs & 3 \\
            Scheduler & Linear  \\ 
    \end{tabular}
    \caption{Hyperparameters for finetuning Gene42 models on GenomicBenchmarks and NT Benchmark.}
    \label{table:Hyperparameter-chromatin-profile-finetune}
\end{table}

\end{document}